\pdfoutput=1
\documentclass[11pt]{article}

\usepackage[final]{template/acl}

\usepackage{times}
\usepackage{latexsym}
\usepackage{verbatim}
\usepackage[T1]{fontenc} 
\usepackage[utf8]{inputenc} 
\usepackage{microtype}
\usepackage{inconsolata}
\usepackage{graphicx} 
\usepackage{array} 
\usepackage{booktabs}
\usepackage{multirow}
\usepackage{tabularx} 
\usepackage{amssymb} 
\usepackage{pifont}  
\usepackage{enumitem} 
\usepackage{longtable}  
\usepackage{float} 

\title{Fane at SemEval-2025 Task 10: \\
Zero-Shot Entity Framing with Large Language Models}

\author{
    Enfa Fane\textsuperscript{*} \quad
    Mihai Surdeanu\textsuperscript{*} \quad
    \textbf{Eduardo Blanco}\textsuperscript{*} \quad
    \textbf{Steven R. Corman}\textsuperscript{†} \\
    \textsuperscript{*}University of Arizona \quad
    \textsuperscript{†}Arizona State University \\
    \textsuperscript{*}\texttt{\{enfageorge, msurdeanu, eduardoblanco\}@arizona.edu} \\
    \textsuperscript{†}\texttt{\{steve.corman\}@asu.edu}
}

\begin{document}
\maketitle
\begin{abstract}
Understanding how news narratives frame entities is crucial for studying media's impact on societal perceptions of events.
In this paper, we evaluate the zero-shot capabilities of large language models (LLMs) in classifying framing roles.
Through systematic experimentation, we assess the effects of input context, prompting strategies, and task decomposition.
Our findings show that a hierarchical approach of first identifying broad roles and then fine-grained roles, outperforms single-step classification.
We also demonstrate that optimal input contexts and prompts vary across task levels, highlighting the need for subtask-specific strategies.
We achieve a Main Role Accuracy of 89.4\% and an Exact Match Ratio of 34.5\%, demonstrating the effectiveness of our approach.
Our findings emphasize the importance of tailored prompt design and input context optimization for improving LLM performance in entity framing.
\end{abstract}
\section{Introduction}\label{sec:introduction}

News is produced and consumed at an unprecedented scale, yet the expectation of neutrality and objectivity in journalism~\citep{doi:10.1177/146488490100200201} often contrasts with the reality that media selectively determines what is newsworthy~\citep{eliaz2024newsmediasuppliersnarratives}.
Through inclusion, omission, or emphasis of specific details, journalists shape public interpretation of events~\citep{iyengar1990framing}, promoting particular recommendations~\citep{entman1993framing}, which in turn influences public opinion and policy decisions~\citep{ziems-yang-2021-protect-serve}.

When such selective framing is used to shape perceptions of an entity, it is known as entity framing~\citep{van-den-berg-etal-2020-doctor}.
SemEval-2025 Task 10 introduces a benchmark for analyzing entity framing~\citep{semeval2025task10,guidelinesSE25T10}, examining how linguistic choices from word selection to narrative structure affect perception.

Example in Figure~\ref{fig:example} demonstrates how strategic language choices portray the entity \textit{Bill Gates} as an Antagonist.
Despite mentioning his substantial investments in climate change initiatives, the narrative emphasizes his high carbon footprint, highlighting hypocrisy.
Phrases such as ``carefully constructed interview'' and references to his connections with Epstein further reinforce an image of deception and evading accountability.
This selective emphasis shapes the reader’s perception of Gates as a Deceiver and Corrupt figure, demonstrating how framing subtly influences interpretation.

\begin{figure}[t]
\centering
\small
\renewcommand{\arraystretch}{1.5}
\begin{tabularx}{\columnwidth}{@{}X@{}}
\toprule
Gates \textit{claimed} that because he continues to spend billions of dollars on climate change activism, his \textit{carbon footprint} isn’t an issue [\dots] Elsewhere during the \textit{carefully constructed interview}, Gates said he was surprised that he was \textit{targeted by `conspiracy theorists' for pushing vaccines} during the pandemic [\dots] While the BBC interview \textit{was set up} to look like Gates was being challenged or grilled, he wasn’t asked about his \textit{close friendship with the elite pedophile Jeffrey Epstein} \\
\bottomrule
\end{tabularx}
\caption{An example document excerpt from the dataset. The author strategically uses loaded language, contrastive framing, and selective emphasis to shape the reader’s perception of Bill Gates as corrupt and deceptive. By highlighting contradictions in his public advocacy, casting doubt on the authenticity of his media appearances, and referencing controversial associations, the text reinforces a negative framing.}\label{fig:example}
\end{figure}

Recent advances in large language models (LLMs) have demonstrated superior performance in text classification, frequently outperforming traditional machine learning approaches on complex tasks~\citep{kostina2025largelanguagemodelstext}.
In this paper, we explore if this extends to entity framing.
Relying solely on prompting, we ask whether LLMs accurately identify the framing roles assigned to entities in news articles.
Given the sensitivity of LLM predictions to prompt variability~\citep{zhuo-etal-2024-prosa}, we assess how prompt engineering strategies such as role-based prompting, incorporating task related information such as label definitions, and rationale generation shape classification performance.

To this end, we develop a modular zero-shot approach that relies exclusively on prompting.
Our method decomposes the task into two stages, first predicting broad narrative roles (Protagonist, Antagonist, Innocent), and then refining into fine-grained roles.
We systematically vary the input context comparing full articles, entity-centered excerpts, and framing-preserving summaries and design prompt templates incorporating expert personas and task related information.

The main contributions of this paper are:
\begin{enumerate}[label={(\arabic*)},itemsep=0.2pt]
\item \textit{Systematic evaluation of LLMs for zero-shot entity framing:} We assess how well LLMs infer framing roles from implicit narrative cues without fine-tuning.

\item \textit{A multi-step prompting strategy:} We show that decomposing the task into two stages, predicting broad roles first, then fine-grained roles, yields better results than a single-step approach.

\item \textit{Strategic prompt design:} We demonstrate that effective prompt design, incorporating role-based guidance and task definitions, improves classification.
A carefully engineered prompt enables a smaller model to perform comparably to a larger one.
\end{enumerate}

We report a Main Role Accuracy of 89.4\% and an Exact Match Ratio of 34.5\% on the SemEval 2025 Task 10 Subtask 1 for English, placing sixth on among the participating teams. We make our codebase and associated prompts publicly available.\footnote{\url{https://github.com/beingenfa/semeval2025task10}}

\section{Background: Task \& Dataset}
\label{sec:background:-task-&-dataset}

The SemEval 2025 Task 10: \textit{Multilingual Characterization and Extraction of Narratives from Online News} introduces three subtasks for analyzing narrative elements in news articles across five languages: Bulgarian, English, Hindi, European Portuguese, and Russian.
This paper focuses on \textit{Subtask 1: Entity Framing} for \textit{English}.

The goal of \textit{Entity Framing} is to determine how a named entity is framed within a news article.
Framing is represented at two levels: \textit{Main Role} and \textit{Fine-Grained Roles}.
The main role broadly categorizes the entity as a Protagonist, Antagonist, or Innocent, while fine-grained roles provide more detailed labels.\footnote{The fine-grained labels are listed in Appendix~\ref{sec:entity-framing-labels}}
An entity may be framed with multiple fine-grained labels, making this a multi-class, multi-label classification problem.

For further details on the dataset, including domain coverage, corpus statistics, and annotation methodology, see~\citep{semeval2025task10,guidelinesSE25T10}.
We now describe our approach to leveraging large language models (LLMs) for entity framing classification.
\section{Related Work}

Language shapes interpretation by selectively emphasizing aspects of reality \citep{entman1993framing}. Prior research identifies two key linguistic mechanisms underlying framing effects: \textit{naming conventions}, where respectful or formal titles correlate strongly with positive sentiment \citep{van-den-berg-etal-2019-president, van-den-berg-etal-2020-doctor}; and \textit{narrative framing}, reflecting partisan differences in media portrayals of events \citep{ziems-yang-2021-protect-serve}. However, these studies primarily treat framing as an implicit phenomenon correlating with other tasks, rather than explicitly classifying or predicting specific framing roles or labels.

Beyond text, in the multimodal domain, \citet{sharma-etal-2022-findings} introduce HVVMemes, a meme dataset covering COVID-19 and US politics,  annotated with coarse-grained framing roles: \textit{hero}, \textit{villain}, \textit{victim}, or \textit{none}. In contrast, our work focuses on  news text and classifies entities at a more fine-grained, hierarchical roles.

Large language models (LLMs) have recently achieved strong performance in text classification tasks, often surpassing traditional machine learning methods on complex datasets~\citep{kostina2025largelanguagemodelstext}. However, significant challenges remain, particularly concerning sensitivity to prompt phrasing and input structure~\citep{zhuo-etal-2024-prosa}. Structured prompting strategies, such as role conditioning~\citep{kong-etal-2024-better} and instruction reframing~\citep{mishra-etal-2022-reframing}, have been proposed. Nevertheless, the effectiveness of these strategies remains highly task-dependent and often requires careful adaptation to specific problem settings~\citep{atreja2024prompt}. 

Guided by these insights, we develop and systematically experiment with multiple prompting strategies for hierarchical entity framing, aiming to improve both coarse and fine-grained entity framing classification.
\section{Prompting for Framing Classification}
\label{sec:prompting-for-framing-classification}

\begin{figure}
    \small
    \centering
    \begin{tabularx}{\columnwidth}{@{}X@{}}
\toprule
\texttt{\{expert\_phrasing\}} \\
You will be provided with \texttt{\{input\_format\_phrase\}}. \\
Your task is to \texttt{\{task\_definition\}}. \\
\texttt{\{task\_instructions\}} \\
\texttt{\{output\_format\_with\_example\}} \\
\texttt{\{input\_context\}} \\
\bottomrule
\end{tabularx}
    \caption{Structure of the prompt template. Curly-bracketed content is dynamically replaced based on the experimental setting
    (see Appendix~\ref{sec:prompt-template-details} for exact phrasing).
    The template is designed to minimize variability, ensuring that observed differences arise solely from targeted prompt modifications.}
    \label{fig:template}
\end{figure}

To systematically evaluate the impact of different prompting strategies on framing classification, we adopt a structured, template-based approach (Figure~\ref{fig:template}).
LLM decision-making is often opaque.
By using a fixed template, we ensure any variation in outputs comes from our controlled prompt modifications rather than randomness.The template consists of modular components that adapt to different experimental conditions.

Our approach is structured around three key components:
\textit{(1)~Input Context} which determines the textual context provided for classification, specified in \texttt{{input\_context}}.
\textit{(2)~Prompt Design}, which examines prompting strategies, such as persona prompting, structured within \texttt{{task\_definition}}, \texttt{{task\_instructions}}, and \texttt{{output\_format\_with\_example}}.
\textit{(3)~Inference Strategy}, which defines whether classification is performed as a single task or decomposed into two steps, reflecting the hierarchical nature of framing.
The following sections provide a detailed discussion of each component.

\subsection{Input Context Variation}
\label{subsec:input-context-variation}

Framing involves selectively emphasizing certain details while omitting others to highlight specific aspects of perceived reality~\citep{entman1993framing}.
Thus, optimizing context is crucial for maximizing narrative signals.
To assess the impact of context granularity on framing classification, we define five input settings encompassing both extractive and summarization-based approaches,\footnote{See Appendix~\ref{sec:summary-generation} for details on summarization.}
each varying in the amount of contextual information available for classification.

\begin{itemize}[noitemsep,topsep=0pt,parsep=0pt,partopsep=0.3pt]
    \item \textit{Full Text} (FT): The entire article.
    \item \textit{Entity-Sentences} (Ent-Sent): Only sentences mentioning the entity.
    \item \textit{Entity-Sentences Neighbors} (Ent-Neigh): Entity-mentioning sentences plus one preceding and one following sentence.
    \item \textit{Neutral Summary} (Neutral-Sum): A summary generated by an LLM prompted to focus neutrally on the entity’s involvement, actions, and framing.
    \item \textit{Framing-Preserved Summary} (FP-Sum): A summary generated by an LLM prompted to preserve the article’s original framing, emphasizing positive or negative actions.
\end{itemize}

\subsection{Prompt Design}
\label{subsec:prompt-design}

While various prompting strategies have been proposed~\citep{gu2023systematic}, we focus on the following:
\begin{enumerate}[label=(\arabic*),noitemsep,topsep=0pt,parsep=0pt,partopsep=0pt]
    \item \textit{Role/Persona Prompting:} Assigning a role or persona in an LLM prompt may shape its performance.
    While one study finds that role-based prompting improves task performance~\citep{kong-etal-2024-better}, another suggests its effectiveness is highly task-dependent and varies across settings~\citep{zheng-etal-2024-helpful}.
    To assess its impact on entity framing, we compare a neutral prompt with one explicitly assigning an expert persona.
    \item \textit{Task-Related Information:} We test two configurations: one providing only the task definition and another including both task and label definitions.
    \item \textit{Output Rationale:} We compare predictions with and without model-generated explanations to assess whether requiring justifications improves or degrades classification accuracy.
\end{enumerate}

\subsection{Inference Strategy}
\label{subsec:inference-strategy}

We compare two inference strategies that determine whether classification is performed as a single task or decomposed into two stages, reflecting the hierarchical structure of framing.
\textit{Single-Step Prediction} jointly predicts both the main and fine-grained roles within a single inference step, whereas
\textit{Multi-Step Prediction} first infers the main role, which is then incorporated into the prompt to predict the fine-grained role.

Having established our prompting strategies, we now present the experimental setup used to systematically evaluate their effectiveness.


\section{Experimental Setup}\label{sec:experimental-setup}

We first discuss a key assumption in our study, followed by evaluation metrics, and LLM setup.

\paragraph{Single-Label Approximation for Fine-Grained Role Classification}
Although the SemEval task defines Entity Framing as a multi-label classification problem, we observe that entity mentions typically receive only one fine-grained role, averaging 1.08 fine-grained roles per mention in the English training split, a trend consistent across languages and main roles (Appendix~\ref{sec:average-sub-roles-across-language}).
Additionally, in our experiments, we observed that the model in our study consistently predicts two main roles even when allowed to predict only one.
Given these patterns, we adopt a single-label approximation, assigning each mention a single fine-grained role.

\paragraph{Evaluation Metrics}
We assess performance using two metrics: \textit{Main Role Accuracy (MRA)}, the proportion of instances where the predicted main role matches the gold label, and \textit{Exact Match Ratio (EMR)}, the proportion of instances where both the main and fine-grained roles match exactly, with no partial credit.
\paragraph{Model and API}
We use GPT-4o (gpt-4o-2024-08-06) via the OpenAI API.\footnote{https://github.com/openai/openai-python, Temperature set to 0 and all other parameters at their default values.
A fixed random seed (42) is specified, though deterministic outputs are not guaranteed.}

We next examine the results obtained using different prompting strategies and input contexts across both inference setups.

\section{Results}\label{sec:results}

We present results from our prompting experiments on the development set, followed by the official SemEval results and post-SemEval analyses.
Key findings are discussed in Section~\ref{sec:insights}.
\subsection{Development}\label{subsec:development}

\begin{table}[t]
    \centering
    \small
    \begin{tabular}{l c c c c}
        \toprule
        \textbf{Metric} & \textbf{Baseline} & \textbf{+ EP} & \textbf{+ LD} & \textbf{+ RA} \\
        \midrule
        \multicolumn{5}{l}{\textbf{Main Role Accuracy}} \\
        FT            & 0.93  & 0.92  & 0.89  & 0.91  \\
        Ent-Sent      & 0.90  & 0.93  & 0.92  & 0.91  \\
        Ent-Neigh     & 0.92  & 0.93  & 0.92  & 0.93  \\
        Neutral-Sum   & 0.69  & 0.73  & 0.69  & 0.71  \\
        FP-Sum    & \textbf{0.95}  & \textbf{0.95}  & 0.93  & 0.93  \\
        \midrule
        \multicolumn{5}{l}{\textbf{Exact Match Ratio}} \\
        FT            & 0.29  & \textbf{0.35}  & 0.30  & 0.29  \\
        Ent-Sent      & 0.32  & 0.33  & \textbf{0.35}  & 0.31  \\
        Ent-Neigh     & 0.30  & 0.34  & 0.32  & 0.31  \\
        Neutral-Sum   & 0.22  & 0.21  & 0.21  & 0.24  \\
        FP-Sum    & 0.33  & 0.32  & 0.34  & 0.33  \\
        \bottomrule
\end{tabular}

      \caption{Performance comparison of different input contexts and prompt engineering strategies when jointly prompting for main role and fine-grained roles.
EP = Expert Persona, LD = Label Definitions, RA = Rationale. For main roles, Framing-Preserved summaries are the strongest input context, achieving the highest accuracy (0.95).
      For fine-grained roles, full text, and only sentences containing entity are strongest when used with a prompting strategy.}
    \label{tab:single-step-prediction-results}

\end{table}

We systematically assess the impact of individual prompt modifications across different input contexts in both single-step and multi-step settings, using the baseline prompt detailed in Appendix~\ref{sec:prompt-template-details}.
To ensure precise attribution of effects, we isolate prompt modifications rather than evaluating combined strategies.
While one combination setting is included in Section~\ref{subsec:official-semeval-results}, systematic evaluation of prompt strategy combinations is beyond the scope of this paper.

\paragraph{Single-Step Approach}

Table~\ref{tab:single-step-prediction-results} shows the performance of different input contexts and prompt engineering strategies in the Single Step Setup.
Main role accuracy is highest with Framing-Preserved Summaries (0.95), where additional prompt strategies yield no further gains.
In contrast, fine-grained role classification benefits from prompting strategies, with Full Text + Expert Persona and Entity-Only Sentences + Label Definitions achieving the highest Exact Match Ratio (0.35).

These findings suggest that a single input strategy may not optimally support both tasks, motivating further exploration of a multi-step approach, where main and fine-grained roles are predicted separately.
Additionally, due to consistently lower performance, the Neutral Summary setting is excluded from subsequent experiments.

\paragraph{Multi-Step Approach}

\begin{table}[t]
    \centering
    \small
    \begin{tabular}{l c c c c}
        \toprule
        \textbf{Metric} & \textbf{Baseline} & \textbf{+ EP} & \textbf{+ LD} & \textbf{+ RA} \\
        \midrule
        FT           & 0.89 & 0.91  & 0.91  & 0.87   \\
        Ent-Sent      & 0.84  & 0.86  & 0.87  & 0.82  \\
        Ent-Neigh     & 0.91  & 0.92  & 0.91  & 0.88  \\
        FP-Sum    & 0.92  & 0.93  & \textbf{0.96}  & 0.93  \\
     \bottomrule
\end{tabular}

    \caption{Performance comparison of different input contexts and prompt engineering strategies in the Multi-Step Setup for Main Role Prediction, reported with accuracy.
EP = Expert Persona, LD = Label Definitions, RA = Rationale.
Framing-Preserved Summaries remain the strongest input context, achieving the highest accuracy (0.96), now benefiting from the Label Definitions (LD) strategy, unlike in the single-step setting.}
   \label{tab:main-role-only-multi-step-prediction}
\end{table}

In the multi-step setting, the model first predicts the main role, which is then used to guide fine-grained classification.

\noindent \textit{Main Role Prediction}: Table~\ref{tab:main-role-only-multi-step-prediction} reports main role accuracy when predicted independently.
Framing-Preserved Summaries remain the strongest setting, achieving 0.96 accuracy.

\begin{table}[t]
    \centering
    \small
    \begin{tabular}{l c c c c}
        \toprule
        \textbf{Metric} & \textbf{Baseline} & \textbf{+ EP} & \textbf{+ LD} & \textbf{+ RA} \\
        \midrule
        FT           & 0.33 & 0.36  & 0.31  & 0.35   \\
        Ent-Sent      & 0.36  & \textbf{0.44}  & 0.35  & 0.37  \\
        Ent-Neigh     & 0.36  & 0.38  & 0.34  & 0.37  \\
        FP-Sum    & 0.29  & 0.29  & 0.35  & 0.31  \\
     \bottomrule
\end{tabular}

    \caption{Performance comparison of different input contexts and prompt engineering strategies in the Multi-Step Setup for fine-grained role classification using Exact Match Ratio metric.
    EP = Expert Persona, LD = Label Definitions, RA = Rationale.
    The multi-step approach improves EMR across most settings compared to single-step prediction, with Entity-Sentences + Expert Persona (EP) achieving the highest EMR (0.44).}
    \label{tab:fine-grained-roles-multi-step-prediction}
\end{table}

\noindent \textit{Fine-grained roles:}
Main role predictions from the best-performing setup are used to inform the prompt for fine-grained role classification.
Additionally, we restrict the set of possible output labels to only those fine-grained roles valid for the predicted main role.
This reduces ambiguity and improves classification accuracy.

As shown in Table~\ref{tab:fine-grained-roles-multi-step-prediction}, the multi-step approach outperforms single-step prediction across most settings.
The best-performing strategy is Entity-Sentences + Expert Persona (EP), which achieves an Exact Match Ratio (EMR) of 0.44, a substantial improvement over the best single-step result (0.35).
Entity-Sentences remains the strongest input context for fine-grained roles(0.44 EMR), whereas full text falls behind.

These findings further support the effectiveness of separating main and fine-grained role prediction and constraining valid role labels to enhance classification accuracy.

\subsection{Official SemEval Results}\label{subsec:official-semeval-results}

\begin{table}[t]
    \centering
    \small
    \setlength{\tabcolsep}{.1in}
       \begin{tabular} {@{}l l r@{\ }r r@{\ }r@{}}
\toprule
\textbf{Rank} & \textbf{Team} & \multicolumn{2}{c}{\textbf{EMR ($\Delta$)}} & \multicolumn{2}{c}{\textbf{MRA ($\Delta$)}} \\
\midrule

1 & DUTIR  & 0.41 &  & 0.95 &   \\

2 & PATeam & 0.38 & {\scriptsize(-0.03)} & 0.89 & {\scriptsize(-0.06)} \\

3 & DEMON & 0.37 & {\scriptsize(-0.04)} & 0.92 & {\scriptsize(-0.03)} \\

4 & gowithnlp & 0.37 & {\scriptsize(-0.04)} & 0.94 & {\scriptsize(-0.01)} \\

5 & TartanTritons & 0.36 & {\scriptsize(-0.05)} & 0.72 & {\scriptsize(-0.23)} \\

6 & \textbf{Ours} & \textbf{0.34} & {\scriptsize\textbf{(-0.07)}} & \textbf{0.89} & {\scriptsize\textbf{(-0.06)}} \\

27 & Baseline & 0.04 & {\scriptsize(-0.37)} & 0.29 & {\scriptsize(-0.66)} \\

\bottomrule
\end{tabular}
    \caption{Official SemEval test set results. Teams are ranked by Exact Match Ratio (EMR), with Main Role Accuracy (MRA) also reported.
Our system ranked 6th, achieving an EMR of 0.34, just 0.07 behind the top system (DUTIR). $\Delta$ values indicate the difference from the top-ranked system.}
    \label{tab:semeval-results}
\end{table}

The findings from our development experiments highlight the effectiveness of different input contexts and prompting strategies.
However, for our official SemEval submission, we employed a Full-Text + Expert Persona + Multi-Step setup using O1(o1-2024-12-17) as our model, as this configuration was selected based on our pre-competition experiments.

As shown in Table~\ref{tab:semeval-results}, our system ranked 6th overall, achieving an Exact Match Ratio (EMR) of 0.34, placing 0.07 behind the top-performing system (DUTIR).
While competitive, our post-SemEval ablation studies revealed a more effective strategy, which we discuss in the next section.

\subsection{Post-SemEval}\label{subsec:post-semeval}

\begin{table}[t]
    \centering
    \footnotesize 
    \setlength{\tabcolsep}{0.07in} 
    \begin{tabular}{lccccc}
        \toprule
        \textbf{Approach} & \textbf{EMR} & \textbf{MRA} & \textbf{Model} & \multicolumn{2}{c}{\textbf{Price/1M Tokens}} \\
        \cmidrule(lr){5-6}
        & & & & \textbf{Input} & \textbf{Output} \\
        \midrule
        SemEval  & 0.345 & 0.894 & O1 & \$15.00 & \$60.00 \\
        Improved & \textbf{0.349} & \textbf{0.894}  & GPT-4o & \$2.50 & \$10.00 \\
        \bottomrule
    \end{tabular}
    \caption{Comparison of our official SemEval submission and the refined approach.
    Although the performance is nearly identical, the refined approach is notable because it achieves these results using GPT-4o, a significantly smaller and more cost-effective model.
    This highlights the importance of optimized prompt design, which allows a more compact model to match or exceed the performance of larger, more expensive alternatives.}
    \label{tab:approach-comparison}
\end{table}

After the SemEval submission, we conducted structured ablation studies to refine our approach, as discussed in Section~\ref{subsec:development}.
As shown in Table~\ref{tab:approach-comparison}, the new approach achieves an EMR of 0.349, compared to 0.345 in our official submission, with MRA remaining unchanged at 0.894.
This result is significant because it was achieved using GPT-4o, a smaller and significantly cheaper model.

\section{Insights}\label{sec:insights}
We present key findings from our experiments, examining the impact of different approaches.

\paragraph{Framing and emphasis: selective highlighting shapes interpretation}
As shown in Table~\ref{tab:single-step-prediction-results}, neutral summaries which present entities in a factual, impartial manner, consistently underperformed compared to framing-preserved summaries.
This aligns with existing framing theory, reinforcing that framing is not only about fact selection but also about selective emphasis to shape interpretation~\citep{entman1993framing}.

\paragraph{Justifications do not improve classification performance}
Our experiments show that requiring models to justify their predictions does not improve classification accuracy.
Performance is primarily determined by the information in the prompt.

\paragraph{More information isn't always better}
Longer contexts aren't always useful.
Main role classification benefits from broad context that establishes overarching narratives, whereas fine-grained roles require focused, entity-specific details.
Excessive input, such as full-text context, can dilute key framing signals, whereas condensed, framing-preserved summaries improve accuracy.
This highlights the importance of tailoring context granularity to the specific classification task.

\paragraph{Multi-Step Approach Improves Performance}
We find that a structured, multi-step classification approach substantially improves performance.
By first predicting the main role and then refining fine-grained classification based on that prediction, the model benefits from clearer context at each stage.
This approach reduces ambiguity and ensures that each classification step is optimized for its specific level of framing.

\paragraph{Prompt Design Enables Small Models to Rival Larger Ones}
Finally, our experiments highlight the effectiveness of carefully engineered prompts.
Carefully crafted prompts allow smaller models to rival larger ones.
A well-optimized prompt for GPT-4o matches or exceeds a less-tuned prompt on the larger, costlier o1 model.
These results suggest that before scaling to larger architectures, improving prompt design for smaller models can yield substantial gains in both efficiency and accuracy.
\section{Conclusion}
\label{sec:conclusion}

In this paper, we explore the effectiveness of large language models (LLMs) for Entity Framing Classification in a zero-shot setting.
While LLMs perform well in broad role classification, fine-grained classification remains challenging.
Our multi-step approach, incorporating distinct input contexts and prompt strategies at each stage, significantly improves overall performance.

\section*{Acknowledgments}
This research was supported by a grant from the U.S.\ Office of Naval Research (N00014-22-1-2596).

\appendix
\clearpage

\label{sec:appendix}
\section{Entity Framing Labels}
\label{sec:entity-framing-labels}

The entity framing taxonomy consists of three main roles and 22 fine-grained roles.
For detailed definitions and annotation guidelines, see~\citep{semeval2025task10,guidelinesSE25T10}.

\begin{itemize}
    \item \textbf{Protagonist:} Guardian, Martyr, Peacemaker, Rebel, Underdog, Virtuous.
    \item \textbf{Antagonist:} Instigator, Conspirator, Tyrant, Foreign Adversary, Traitor, Spy, Saboteur, Corrupt, Incompetent, Terrorist, Deceiver, Bigot.
    \item \textbf{Innocent:} Forgotten, Exploited, Victim, Scapegoat.
\end{itemize}
\section{Average Sub Roles across Language}\label{sec:average-sub-roles-across-language}

\begin{table}[h]
\centering
\begin{tabular}{lccccc}
        \toprule
        & \textbf{EN} & \textbf{BG} & \textbf{PT} & \textbf{HI} & \textbf{RU} \\
        \midrule
        \textbf{All}         & 1.08 & 1.13 & 1.05 & 1.16 & 1.06 \\
        \textbf{Protagonist} & 1.06 & 1.02 & 1.01 & 1.24 & 1.00 \\
        \textbf{Antagonist}  & 1.10 & 1.19 & 1.09 & 1.12 & 1.11 \\
        \textbf{Innocent}    & 1.02 & 1.01 & 1.00 & 1.06 & 1.01 \\
        \bottomrule
\end{tabular}

\caption{The average number of sub-roles per instance across different languages (English (EN), Bulgarian(BG), European Portuguese(PT), Hindi(HI),  and Russian(RU)) and main roles in the training split.
The values remain close to one, justifying our single-label approximation for fine-grained classification.}
\label{tab:subrole_distribution}
\end{table}
\onecolumn

\section{Prompts for Summary Generation}
\label{sec:summary-generation}

\subsection{Neutral Summary}\label{subsec:neutral-summary}

\begin{figure*}[!htpb]
\small
\centering
\begin{tabularx}{\linewidth}{@{}X@{}}
\toprule
\texttt{Summarize the following article with a specific focus on \{ entity \}.
Write the summary as a standalone description, ensuring that the entity and its role are clearly introduced without referring to 'the article' or assuming prior context.
Clearly state their involvement, actions, and framing within the event.
Maintain a factual and neutral tone.}\\
\bottomrule
\end{tabularx}
\caption{Prompt used to generate neutral summaries. The instructions guide the LLM to ensure that that the entity’s role is explicitly introduced while maintaining a factual and impartial tone.}
\label{fig:neutral-prompt}
\end{figure*}

\subsection{Framing-Preserved Summary}\label{subsec:framing-preserved-summary}

\begin{table*}[ht]
    \centering
    \small
    \renewcommand{\arraystretch}{1.2}
\begin{tabularx}{\linewidth}{@{}X@{}}
\toprule
\texttt{Write a standalone summary that clearly reflects how the author of the article frames {entity} and their actions.
Do not present the entity neutrally—mirror the language and implicit bias of the article itself.
If the author portrays the entity favorably, highlight their positive actions, successes, and beneficial impact.
If the author is critical, emphasize the entity’s negative actions, failures, or harmful consequences.}\\

\texttt{If the framing is mixed or subtle, encode the contrast in tone and nuance.Strongly reflect the author’s framing through:}\\
\texttt{- Loaded or emotionally charged language (if present in the article)}\\
\texttt{- Emphasis on the entity’s perceived intentions and motivations}\\
\texttt{- Who is affected by their actions and how the consequences are framed}\\
\texttt{- Any direct or implied judgments made by the author.}\\
\texttt{Maintain the style and tone of the article, ensuring the framing is explicit in how the entity’s role and impact are described. Do not add external information or neutralize the bias. The summary should feel as though it was written by the original author.}\\
\bottomrule
\end{tabularx}
\caption{Prompt used to generate framing-preserved summaries.
The instructions guide the model to reflect the article’s original framing, emphasizing bias and tone while avoiding neutrality or external information.}
\label{fig:framing-preserved summary}
\end{table*}

\section{Prompt Template Details} \label{sec:prompt-template-details}

Our system prompt follows the structured format introduced earlier.
The input context is provided in the user prompt, while the rest remains part of the system prompt.

\begin{table}[H]
    \centering
    \small
    \renewcommand{\arraystretch}{1}
    \begin{tabularx}{0.7\linewidth}{@{} p{3cm} X@{}}
    \toprule
    \textbf{Prompt Type} & \textbf{Template} \\
    \midrule
    \textbf{System Prompt} & \texttt{\{expert\_phrasing\}} \\
    & You will be provided with \texttt{\{input\_format\_phrase\}}. \\
    & Your task is to \texttt{\{task\_definition\}}. \\
    & \texttt{\{task\_instructions\}} \\
    & \texttt{\{output\_format\_with\_example\}} \\\\
    \textbf{User Prompt} & \texttt{\{input\_context\}} \\
    \bottomrule
    \end{tabularx}
    \caption{Structure of the system and user prompts. The system prompt provides instructions, while the user prompt supplies contextual input for the model.}
    \label{tab:structure}
\end{table}

We outline the values these elements take in different settings below.
\subsection{Input Format Phrase \& Input Context}\label{subsec:input-format-phrase-&-input-context}

The input context is formatted as \texttt{DOCUMENT:\{content\}} Note that in the \texttt{Entity-Sentences} and \texttt{Entity-Sentences Neighbors} settings, the individual sentences or sentence groups are separated by \texttt{[...]}.

\noindent  In the system prompt, the input format phrase specifies the type of input context provided to the model.
The format follows this structure:\texttt{a \{document\_type\_str\} in the following format- DOCUMENT:\{document\_type\_str\}}.
The corresponding \texttt{document\_type\_str} for each input context type is provided in the table below:

\begin{table}[H]
    \centering
    \small
    \renewcommand{\arraystretch}{1.2}
    \begin{tabularx}{\linewidth}{@{}X X@{}}
    \toprule
    \textbf{Input Context Type} & \textbf{\texttt{document\_type\_str}} \\
    \midrule
    Full Text & \texttt{news article} \\
    Entity-Sentences or Entity-Sentences Neighbors & \texttt{excerpt of a news article} \\
    Neutral Summary or Framing-Preserved Summary & \texttt{news article summary}\\
    \bottomrule
    \end{tabularx}
    \caption{Mapping between input context types and corresponding input phrases.}
    \label{tab:input-format-phrase}
\end{table}
\subsection{Prompt Design}
\label{subsec:prompt-design2}

\subsubsection{Expert Persona}

In the context of role/persona prompting, if no specific persona is assigned, \texttt{\{expert\_phrasing\}} remains empty.
Otherwise, it is replaced with the following text:
\begin{quote}
    \small
    \texttt{You are an expert in analyzing how a specific named entity is portrayed in a given text. Read the text carefully and focus on everything said about \{entity\}.}
\end{quote}
\subsubsection{Output Rationale}

In the template below, we incorporate prompting for justification by including three variables:
\begin{itemize}
    \item \texttt{\{ask\_reasoning\}}: \texttt{"with a reasoning for your prediction"}
    \item \texttt{\{reasoning\_in\_json\}}: \texttt{', "reasoning" : "your reasoning here" }
    \item \texttt{\{reasoning\_example\}}: \texttt{, "reasoning" : "The article frames Kremlin propagandists as instigators and deceivers, highlighting their role in spreading falsehoods and promoting extreme measures."}
\end{itemize}

When the setting does not require justification with the output label, these variables remain empty.

\subsubsection{Task Related Information}

\begin{table}[H]
    \centering
    \tiny
    \renewcommand{\arraystretch}{1.2}
    \raggedright
\begin{tabularx}{\linewidth}{@{} p{1.5cm} p{3.5cm} p{5cm} p{5cm} @{}}
    \toprule
    \textbf{Setting} & \textbf{\texttt{task\_definition}} & \textbf{\texttt{task\_instructions}} & \textbf{\texttt{output\_format\_with\_example}} \\
    \midrule
    Single Step & \texttt{classify the narrative framing of the \{entity\} in the document based on the taxonomy that follows in the format [(broad role,[list of valid fine grained roles/])]. The taxonomy is \{taxonomy\}.} & \texttt{Instructions: 1. Assign exactly one broad role from: Protagonist, Antagonist, or Innocent. 2. Determine one or a maximum of two corresponding fine grained role from the taxonomy. 3. Order the sub-roles by likelihood, with the most likely fine-grained role listed first.\{label\_definitions\} } & \texttt{4. Finally, return your conclusion \{ask\_reasoning\} as a single JSON object with no extra text, in this format:
    \{ "main\_role": "<most\_likely\_main\_role>", "fine\_grained\_roles": ["<Most likely sub-role>",
        "<Second most likely sub-role if relevant>"] \{reasoning\_in\_json\}\}
        Example Output: \{"main\_role": "Antagonist", "fine\_grained\_roles": ["Conspirator"] \{reasoning\_example\}} \\
\midrule
    Multi-Step : Main Role & \texttt{classify the narrative framing of the {entity} in the document as either Protagonist, Antagonist or Innocent.} & \texttt{Instructions: 1. Assign exactly one main role from: Protagonist, Antagonist, or Innocent.\{label\_definitions\}} &
    \texttt{2.Return your conclusion \{ask\_reasoning\} as a single JSON object with no extra text, in this JSON format: \{ "main\_role": "<most\_likely\_main\_role>" \{reasoning\_in\_json\}\} Example Output: \{"main\_role": "Antagonist" \{reasoning\_example\}\}} \\
    \midrule
    Multi-Step : Fine-grained Role & \texttt{classify the narrative framing of the \{entity\} in the document. The taxonomy is \{taxonomy\}. \{label\_definitions\}} & \texttt{You have previously identified the broader narrative frame to be {main\_role\_candidate}. Instructions: 1. Determine one or a maximum of two corresponding fine grained role from the taxonomy. 2. Order the sub-roles by likelihood, with the most likely fine-grained role listed first.} & \texttt{3. Finally, return your conclusion \{ask\_reasoning\} as a single JSON object with no extra text, in this format:
\{ "fine\_grained\_roles": ["<Most likely sub-role>", "<Second most likely sub-role if relevant>"] \{reasoning\_in\_json\}\}
Example Output: \{"fine\_grained\_roles": ["Conspirator"] \{reasoning\_example\} \}} \\
    \bottomrule
    \end{tabularx}
    \caption{Mapping between different settings, task definitions, and corresponding instructions for narrative framing classification.}
    \label{tab:task-definition-instructions}
\end{table}


\paragraph{Taxonomy}
\label{subsec:taxonomy}

The taxonomy shared in the prompt is the same as in Appendix~\ref{sec:entity-framing-labels}.
The label definitions provided are from the official paper and is shared in Table~\ref{tab:task-definition-instruction}.

\begin{table}[H]
    \centering
    \tiny
    \renewcommand{\arraystretch}{1.5}
\begin{tabularx}{\linewidth}{@{} p{1cm} X @{}}
        \toprule
        \textbf{Role Level} & \textbf{\texttt{\{label\_definitions\}}} \\
        \midrule
        Main Role &
         \texttt{Protagonist: The central figure or party in a news article, often portrayed as a hero or positive force driving the narrative.}

         \texttt{Antagonist: The opposing figure or force in a news article, often depicted as the source of conflict or challenge to the protagonist.}

         \texttt{Innocent: An individual or group portrayed as untainted or blameless in the context of the news, typically victimized or wronged.} \\
        Fine Grained Roles Protagonist & \texttt{Guardians: Heroes or guardians who protect values or communities, ensuring safety and upholding justice. They often take on roles such as law enforcement officers, soldiers, or community leaders (e.g., climate change advocacy community leaders).
        Martyr: Martyrs or saviors who sacrifice their well being, or even their lives, for a greater good or cause. These individuals are often celebrated for their selflessness and dedication. This is mostly in politics, not in Climate Change.
        Peacemaker: Individuals who advocate for harmony, working tirelessly to resolve conflicts and bring about peace. They often engage in diplomacy, negotiations, and mediation. This is mostly in politics, not in Climate Change.
        Rebel: Rebels, revolutionaries, or freedom fighters who challenge the status quo and fight for significant change or liberation from oppression. They are often seen as champions of justice and freedom.
        Underdog: Entities who are considered unlikely to succeed due to their disadvantaged position but strive against greater forces and obstacles. Their stories often inspire others.
        Virtuous: Individuals portrayed as virtuous, righteous, or noble, who are seen as fair, just, and upholding high moral standards. They are often role models and figures of integrity. } \\
        Fine Grained Roles Antagonist & \texttt{Conspirator: Those involved in plots and secret plans, often working behind the scenes to undermine or deceive others. They engage in covert activities to achieve their goals.
        Instigator: Individuals or groups initiating conflict, often seen as the primary cause of tension and discord. They may provoke violence or unrest.
        Deceiver: Deceivers, manipulators, or propagandists who twist the truth, spread misinformation, and manipulate public perception for their own benefit. They undermine trust and truth.
        Incompetent: Entities causing harm through ignorance, lack of skill, or incompetence. This includes people committing foolish acts or making poor decisions due to lack of understanding or expertise. Their actions, often unintentional, result in significant negative consequences.
        Corrupt: Individuals or entities that engage in unethical or illegal activities for personal gain, prioritizing profit or power over ethics. This includes corrupt politicians, business leaders, and officials.
        Tyrant: Tyrants and corrupt officials who abuse their power, ruling unjustly and oppressing those under their control. They are often characterized by their authoritarian rule and exploitation.
        Foreign Adversary: Entities from other nations or regions creating geopolitical tension and acting against the interests of another country. They are often depicted as threats to national security. This is mostly in politics, not in Climate Change.
        Terrorist: Terrorists, mercenaries, insurgents, fanatics, or extremists engaging in violence and terror to further ideological ends, often targeting civilians. They are viewed as significant threats to peace and security. This is mostly in politics, not in Climate Change.
        Bigot: Individuals accused of hostility or discrimination against specific groups. This includes entities committing acts falling under racism, sexism, homophobia, Antisemitism, Islamophobia, or any kind of hate speech. This is mostly in politics, not in Climate Change.
        Saboteur: Saboteurs who deliberately damage or obstruct systems, processes, or organizations to cause disruption or failure. They aim to weaken or destroy targets from within.
        Traitor: Individuals who betray a cause or country, often seen as disloyal and treacherous. Their actions are viewed as a significant breach of trust. This is mostly in politics, not in Climate Change.
        Spy: Spies or double agents accused of espionage, gathering and transmitting sensitive information to a rival or enemy. They operate in secrecy and deception. This is mostly in politics, not in Climate Change. } \\
        Fine Grained Roles Innocent & \texttt{Victim: People cast as victims due to circumstances beyond their control, specifically in two categories: (1) victims of physical harm, including natural disasters, acts of war, terrorism, mugging, physical assault, etc., and (2) victims of economic harm, such as sanctions, blockades, and boycotts. Their experiences evoke sympathy and calls for justice, focusing on either physical or economic suffering.
Scapegoat: Entities blamed unjustly for problems or failures, often to divert attention from the real causes or culprits. They are made to bear the brunt of criticism and punishment without just cause.
Exploited: Individuals or groups used for others’ gain, often without their consent and with significant detriment to their wellbeing. They are often victims of labor exploitation, trafficking, or economic manipulation. Forgotten: Marginalized or overlooked groups who are often ignored by society and do not receive the attention or support they need. This includes refugees, who face systemic neglect and exclusion.} \\
        \bottomrule
    \end{tabularx}
    \caption{Label Definitions shared in the prompt. These are from the task definitions~\citep{semeval2025task10}.}
    \label{tab:task-definition-instruction}
\end{table}

\end{document}